\documentclass[sigconf]{acmart}
\usepackage{multirow}
\usepackage{bm}
\usepackage{xspace}
\usepackage{balance}
\usepackage{amsmath}

\newcommand{\eat}[1]{}
\newcommand{\paratitle}[1]{\vspace{1ex}\noindent \textbf{#1}}
\let\oldhat\hat
\renewcommand{\vec}[1]{\bm{#1}}
\renewcommand{\hat}[1]{\oldhat{\mathbf{#1}}}
\renewcommand{\matrix}[1]{\bm{#1}}
\newcommand{\eg}{\emph{e.g., }\xspace}
\newcommand{\rf}{\emph{rf. }\xspace}

\newcommand{\ie}{\emph{i.e., }\xspace}

\newcommand{\etal}{\emph{et al. }\xspace}

\newcommand\BibTeX{B\textsc{ib}\TeX}

\def\BibTeX{{\rm B\kern-.05em{\sc i\kern-.025em b}\kern-.08emT\kern-.1667em\lower.7ex\hbox{E}\kern-.125emX}}

\copyrightyear{2019}
\acmYear{2019}
\acmConference[CIKM '19]{The 28th ACM International Conference on Information and Knowledge Management}{November 3--7, 2019}{Beijing, China}
\acmBooktitle{The 28th ACM International Conference on Information and Knowledge Management (CIKM '19), November 3--7, 2019, Beijing, China}
\acmPrice{15.00}
\acmDOI{10.1145/3357384.3357937}
\acmISBN{978-1-4503-6976-3/19/11}

\begin{document}

\fancyhead{}

\title{Emotion-aware Chat Machine: Automatic Emotional Response Generation for Human-like Emotional Interaction}

\author{
Wei Wei\textsuperscript{1},
Jiayi Liu\textsuperscript{1},
Xianling Mao\textsuperscript{2},
Guibing Guo\textsuperscript{3},
Feida Zhu\textsuperscript{4},
Pan Zhou\textsuperscript{5},
Yuchong Hu\textsuperscript{6},
}
\affiliation{\textsuperscript{1} Cognitive Computing and Intelligent Information Processing (CCIIP) Laboratory, School of Computer Science and Technology, Huazhong University of Science and Technology}
\affiliation{\textsuperscript{2} School of Computer Science and Technology, Beijing Institute of Technology}
\affiliation{\textsuperscript{3} Software College, Northeastern University}
\affiliation{\textsuperscript{4} School of Information Systems, Singapore Management University}
\affiliation{\textsuperscript{5} School of Electronic Information and Communications, Huazhong University of Science and Technology}
\affiliation{\textsuperscript{6} School of Computer Science and Technology, Huazhong University of Science and Technology}
\affiliation{\textsuperscript{1} \{weiw, liujiayi7, panzhou, yuchonghu\}@hust.edu.cn\quad\textsuperscript{2} maoxl@bit.edu.cn\quad\\
\textsuperscript{3} guogb@swc.neu.edu.cn\quad\textsuperscript{4} fdzhu@smu.edu.sg}

\renewcommand{\shortauthors}{Trovato and Tobin, et al.}

\begin{abstract}
The consistency of a response to a given post at \textbf{\emph{semantic}}-level and \textbf{\emph{emotional}}-level is essential for a dialogue system to deliver human-like interactions. However, this challenge is not well addressed in the literature, since most of the approaches neglect the emotional information conveyed by a post while generating responses. This article addresses this problem by proposing a \emph{unified} end-to-end neural architecture, which is capable of simultaneously encoding the \emph{semantics} and the \emph{emotions} in a post for generating more intelligent responses with appropriately expressed emotions. Extensive experiments on real-world data demonstrate that the proposed method outperforms the state-of-the-art methods in terms of both content coherence and emotion appropriateness.
\end{abstract}

%


\settopmatter{printacmref=false, printfolios=false}

\maketitle

{\fontsize{8pt}{8pt} \selectfont
\textbf{ACM Reference Format:}\\
Wei Wei, Jiayi Liu, Xianling Mao, Guibing Guo, Feida Zhu, Pan Zhou, Yuchong Hu. 2019. Emotion-aware Chat Machine: Automatic Emotional Response Generation for Human-like Emotional Interaction. In {\it The 28th ACM International Conference on Information and Knowledge Management (CIKM’19), November 3--7, 2019, Beijing, China.} ACM, New York, NY, USA, 10 pages. https://doi.org/10.1145/3357384.3357937 }

\section{Introduction}
\label{sec:intro}
Dialogue systems in practice are typically built for various purposes like emotional interaction, customer service or information acquisition, which can be roughly categorized into three classes, \ie chitchat chatbots, task-oriented chatbots and domain-specific chatbots.
For example,
a task-specific chatbot can serve as a customer consultant,
while a chitchat chatbot is commonly designed for convincingly simulating how a human would respond as a conversational partner.
In fact, most recent work on response generation in chitchat domain can be summarized as follows, \ie retrieval-based, matching-based, or statistical machine learning based approaches~\cite{wallace2003elements,wilcox2011beyond,ritter2011data,ji2014information}.
Recently, with the increasing popularity of deep learning, many research efforts have been dedicated to employing an encode-decoder architecture~\ie Sequence-to-sequence (\emph{Seq2seq}) models ~\cite{sutskever2014sequence,cho2014learning}, to map a post to the corresponding response with little hand-crafted features or domain-specific knowledge for the conversation generation problem~\cite{shang2015neural,vinyals2015neural}. Subsequently, several variants of \emph{Seq2seq} models are also proposed to address different issues~\cite{serban2016building,xing2017topic,qian2017assigning,mou2016sequence}.

Despite the great progress made in neural dialogue generation, a general fact is that few work has been reported to automatically incorporate emotional factors for dialogue systems. In fact, several empirical studies have proven that chatbots with the ability of emotional communication with humans are essential for  enhancing user satisfaction~\cite{martinovski2003breakdown,prendinger2005using,callejas2011predicting}. To this end,  it is highly valuable and desirable to develop an emotion-aware chatbot that is capable of perceiving/expressing emotions and emotionally interacting with the interlocutors. In literature, Zhou \etal \cite{zhou2018emotional} successfully build an emotional chat machine (ECM) that is capable of generating emotional responses according to a pre-defined emotion category, and several similar efforts are also made by \cite{peng2019topic,huang2018automatic}, such as \cite{zhou2017mojitalk} proposed by Zhou~\etal that utilizes emojis to control the emotional response generation process within conditional variational auto-encoder (CAVE) framework.

\begin{figure}[!t]
\centering
\includegraphics[scale=0.5]{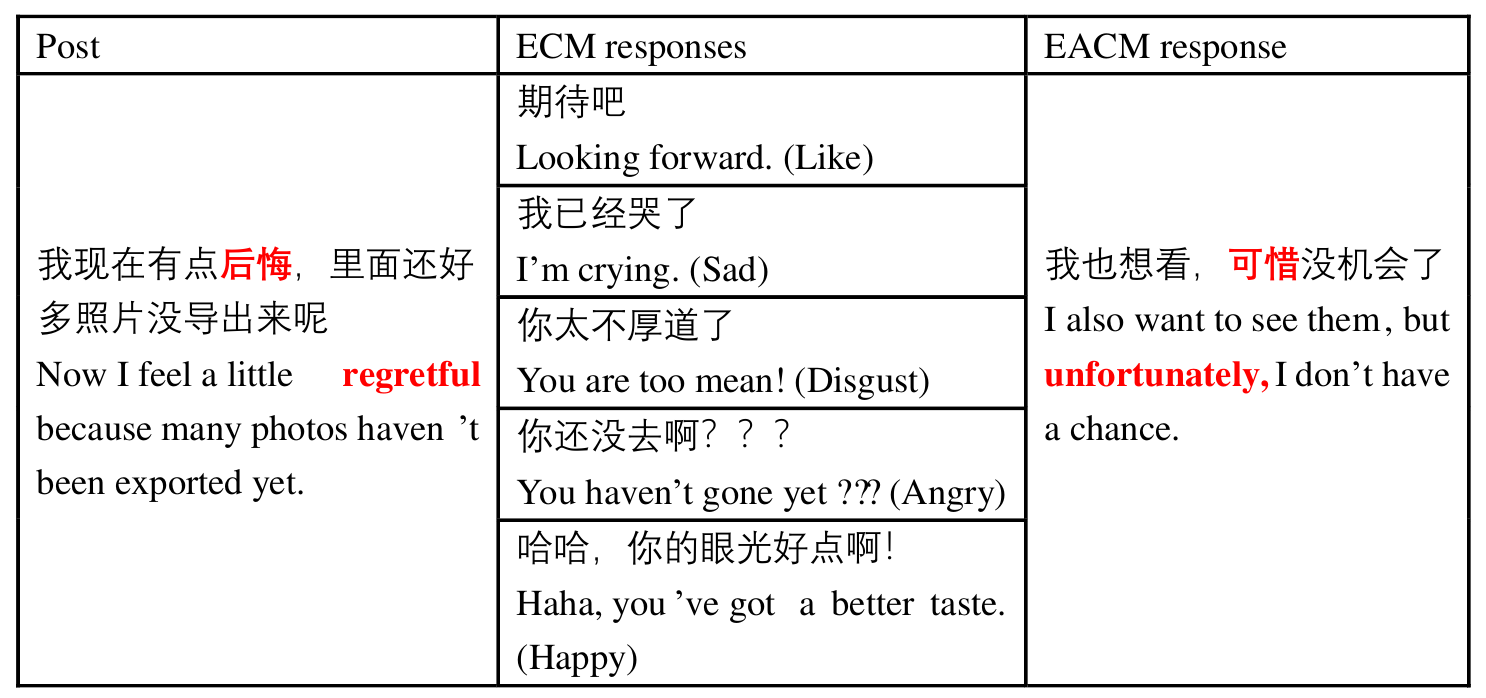}
\caption{Sampled dialogue generated from ECM and EACM.}
\label{fig1}
\vspace{-0.5cm}
\end{figure}

Nevertheless, these mentioned approaches cannot work well owing to the following facts:
(i) These approaches solely generate the emotional responses based on a pre-specified label (or emoji) as shown in Figure~\ref{fig1}, which is unrealistic in practice as the well-designed dialogue systems need to wait for a manually selected emotion category for response generation;
(ii) The generation process apparently divided into two parts would significantly reduce the smoothness and quality of generating responses;
and (iii) As shown in Figure~\ref{fig1},  some emotionally-inappropriate responses (even conflicts) might apparently affect the interlocutor's satisfaction. Thereby, here we argue that  fully exploiting the  emotional information of the given post to supervise the learning process is definitely beneficial for automatically generating responses with the optimal emotion~(\rf ``EACM response'' generated by our method shown in Figure~~\ref{fig1}).

Previous methods greatly contribute to emotion-aware conversation generation problem, however, they are insufficient and several issues emerge when trying to fully-address this problem.
\textbf{\emph{First}}, it is not easy to model human emotion from a given sentence due to semantic sparsity. Psychological studies~\cite{plutchik1980general,plutchik2001nature} demonstrate that human emotion is quite complex and one sentence may contain multi-types of emotions with different intensities.
For example, a hotel guest might write a comment like ``The environment is not bad, however the location is too remote.''
As such, solely using the post's emotion label is insufficient and we need to appropriately extract the emotional information of the input post for representation.
\textbf{\emph{Second}}, it is difficult for a model to decide the optimal response emotion for generation, and it is also not reasonable to directly map the post's emotion label to the response's emotion label, as the emotion selection process is determined not only by the post's emotion but also by its semantic meaning.
\textbf{\emph{Third}}, it is also problematic to design a unified model that can generate plausible emotional sentence without sacrificing grammatical fluency and semantic coherence~\cite{zhou2018emotional}.
Hence, the response generation problem faces a significant challenge: that is,
how to effectively leverage the emotion and semantic of a given post to automatically learn the emotion interaction mode for emotion-aware response generation within a unified model.

In this paper, we propose a novel emotion-aware chat machine (\ie \textsf{EACM} for short), which is capable of perceiving other interlocutor's feeling (\ie post’s emotion) and generating plausible response with the optimal emotion category (\ie response's emotion).
Specifically, \textsf{EACM} is based on a unified \emph{Seq2seq} architecture with a \emph{self-attention} enhanced emotion selector and an emotion-biased response generator, to simultaneously modeling the post's emotional and semantic information for automatically generating appropriate response.
Experiments on the public datasets demonstrate the effectiveness of our proposed method, in terms of two different types of evaluation metrics, \ie \emph{automatic metric} and\emph{ human evaluation}, which are used to measure the diversity of words in the generated sentences (\emph{automatic metric}, indirectly reflecting the diversity of expressed emotions, \eg {\it distinct-n}), and whether the generated responses' emotions are appropriate according to human annotations~(\emph{human evaluation}). The main contributions of this research are summarized as follows:

\begin{enumerate}
    \item It advances the current emotion conversation generation problem from a new perspective, namely emotion-aware response generation, by taking account of the emotion interactions between interlocutors.
    \item It also proposes an innovative generation model (\ie \textsf{EACM}) that is capable of extracting post's sentimental and semantic information for generating intelligible responses with appropriate emotions.
    \item It conducts extensive experiments on a real-word dataset, which demonstrate the proposed approach outperforms the state-of-the-art methods at both \emph{semantic}-level and \emph{emotional}-level.
\end{enumerate}

\section{Related Works}
\label{sec:r_work}
The current conversation generation approaches are mostly based on the basic Sequence-to-sequence (\emph{Seq2seq}) framework, which has been proven~\cite{sutskever2014sequence,cho2014learning} that is able to effectively address the sequence-mapping issues.
Following the success of \emph{Seq2seq} model,  methods based on such framework have been applied to various domains, such as machine translation~\cite{bahdanau2014neural} and image caption generation~\cite{anderson2018bottom}.

Indeed, there exist some attempts on improving the performance of such encoder-decoder architecture for machine translation problem.
Bahdanau \etal~\cite{bahdanau2014neural} utilize Bi-directional Long Short-Term Memory (Bi-LSTM) network with attention mechanism
for long-sentence generation, which is able to automatically search for relevant parts in the context.
Luong~\etal \cite{luong2015effective} thoroughly evaluate the effectiveness of different types of attention mechanisms, \ie global/local attention with different alignment functions.
Furthermore, self-attention mechanism proposed by~\cite{lin2017structured,vaswani2017attention}
is proved effective for machine translation, which can yield large gains in terms of  BLEU~\cite{papineni2002bleu} as compared to the state-of-the-art methods.
To cope with the increasing complexity in the decoding stage (caused by large-scale vocabularies),
Jean~\etal~\cite{jean2014using} consider to use sampled softmax methods and thus achieve encouraging results.
These works have improved the generation performance of the \emph{Seq2seq} model and speeded up decoding process, which build a solid foundation for the future studies based on this architecture.

There also exist many efforts dedicated to research on how to apply Seq2seq model for conversation systems
~\cite{shang2015neural,sordoni2015neural,vinyals2015neural}, by regarding the conversation generation as a \emph{monolingual translation} task,
and later on several variants are proposed for a wide variety of domain-specific issues,
such as
hierarchical recurrent model~\cite{serban2016building,serban2017hierarchical}, topic-aware model~\cite{xing2017topic}.
Besides, several persona-based~\cite{li2016persona} models and identity-coherent models~\cite{qian2017assigning} are proposed to endow the chatbots with personality
for addressing the context-consistency problem.
There have been numerous attempts to generate more diverse and informative responses, such as
Maximum Mutual Information (MMI) based model~\cite{li2015diversity} and enhanced beam-search based model~\cite{li2016simple,vijayakumar2016diverse}.
Several approaches are also proposed for specific tasks, such as Zhou~\etal~\cite{zhou2018commonsense} take account of static graph attention to incorporate commonsense knowledge for chatbots.
Zhang~\etal~\cite{zhang2018tailored} propose different solutions for two classical conversation scenarios, \ie chit-chat and domain-specific conversation.

In recent years, many researches propose that emotion factors are of great significance in terms of successfully building human-like conversation generation models.
Ghosh~\etal~\cite{ghosh2017affect} propose \emph{affect language model} to generate texture conditioned on the given affect categories with controllable affect strength.
Hu~\etal~\cite{hu2017toward} present a combined model of the Variational Auto-Encoder (VAE) and holistic attribute discriminators to generate sentences with certain types of sentiment and tense. However, these models are mainly built for emotional text generation task.
Several proposals study the conversation generation problem with emotional factors, which are most related to our proposed conversation generation problem.
Zhou~\etal~\cite{zhou2018emotional} develop an Emotional Chat Machine (ECM) model using three different mechanisms (\ie \emph{emotion embedding}, \emph{internal memory} and \emph{external memory}) to generate responses according to the designated emotion category.
Similarly, Zhou~\etal~\cite{zhou2017mojitalk} propose a reinforcement learning approach within conditional variational auto-encoder framework to generate responses conditioned on the given emojis.
In ~\cite{huang2018automatic}, Huang~\etal propose three different models that are capable of injecting different emotion factors for response generation.
Peng~\etal~\cite{peng2019topic} utilize Latent Dirichlet allocation (LDA) models to extract topic information for emotional conversation generation.
However, all of such models are based on a designated emotion category to generation emotional responses, which need human beings to decide an optimal response emotion category for generation. Besides, the emotion information of the given post is not explicitly modeled, and thus the generated responses are not good enough.
As opposed, our proposed model is capable of effectively leverage the emotion and semantics of a given post to automatically learn the emotion interaction mode for
emotion-aware response generation.

\section{Proposed Model}
\label{sec:model}

\begin{figure*}[!t]
\centering
\includegraphics[scale=0.4]{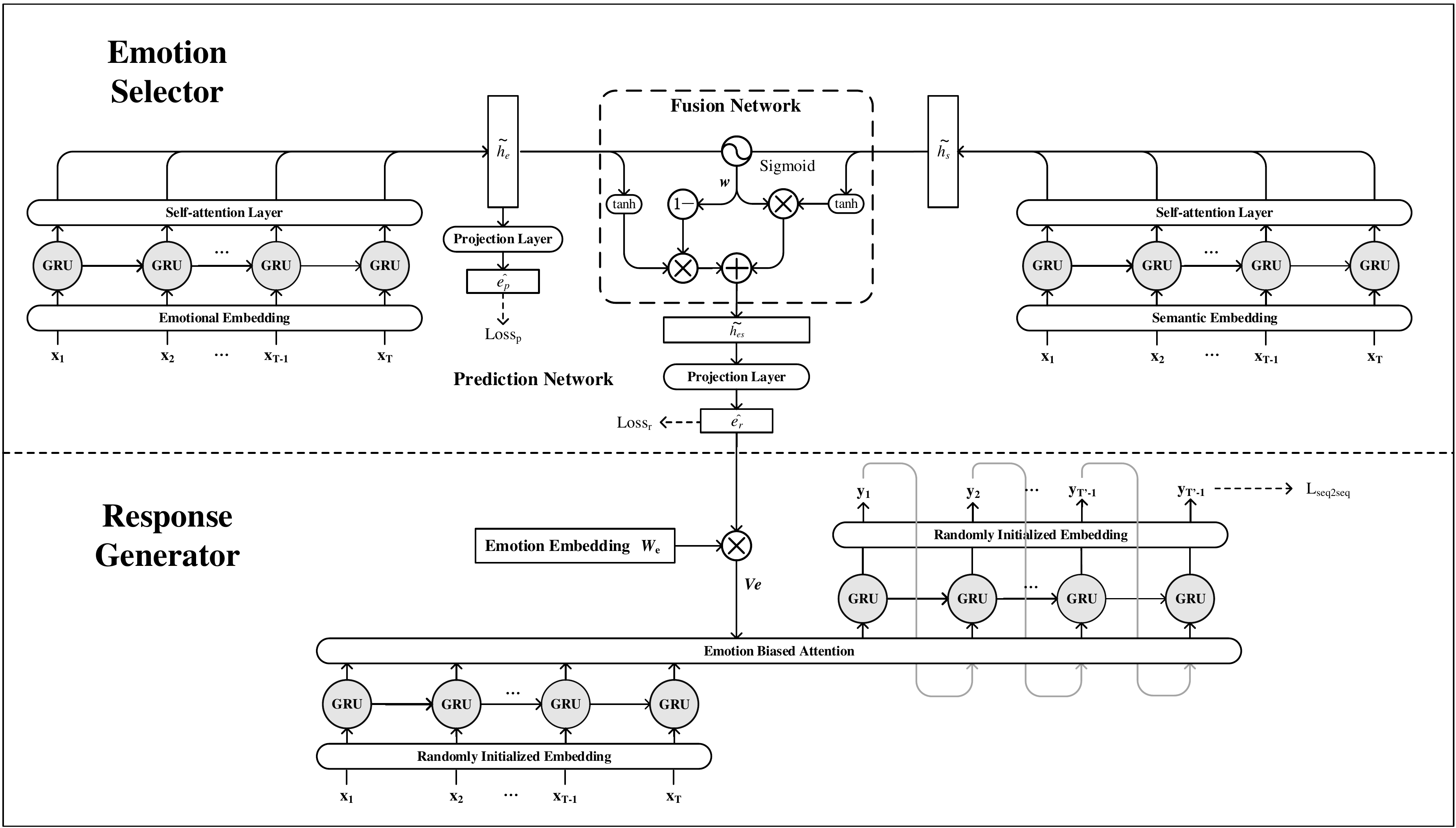}
\caption{Overview of EACM.}
\label{fig2}
\end{figure*}

\subsection{Preliminary: Sequence-to-Sequence Attention Model}
In the literature, sequence-to-sequence (\emph{Seq2seq}) model is widely adopted for dialogue generation~\cite{bahdanau2014neural,cho2014learning,sutskever2014sequence}.
In order to promote the quality of the generated sentences, the Seq2seq-attention model~\cite{bahdanau2014neural} is proposed for dynamically
attending on the key information of the input post during decoding.
In this paper, our approach is mainly based on \emph{Seq2seq-attention} models for response generation and therefore we will first illustrate this basic model in principle.

The Seq2seq-attention model is typically a deep RNN-based architecture with an encoder and a decoder. The encoder takes the given post sequence  $\vec{x}= \{x_1,x_2,\cdots, x_{T}\}$~($T$ is the length of the post) as inputs, and maps them into hidden representations $\vec{h} = (\vec{h}_1,\vec{h}_2,\cdots, \vec{h}_{T})$.  The decoder then decodes them to generate a possibly variably-sized word sequence, \ie $\vec{y} = \{y_1, y_2, \cdots, y_{T^{'}}\}$, where $T^{'}$ is the length of the output sequence, and it may differ from $T$.

In more detail, the context representation $\vec{c}_{t}$ of the post sequence $\vec{x}$ is computed by parameterizing the encoder hidden vector $\vec{h}$ with different attention scores~\cite{bahdanau2014neural}, that is,
\begin{equation}
  \vec{c}_{t}=\sum_{j=1}^{T}\vec{\alpha}(\vec{s}_{t-1},\vec{h_j})\cdot \vec{h_j},
\end{equation}
where $\vec{\alpha}(.,.)$ is a coefficient estimated by each encoder token's relevance to the predicting $y_t$.
The decoder iteratively updates its state $\vec{s}_t$ using previously the generated word $\vec{y}_{t-1}$, namely,
\begin{equation}
\label{eq:over_opt}
  \vec{s}_t = f(\vec{s}_{t-1},\vec{y}_{t-1},\vec{c_t}),\quad\quad t  = 1, 2, \cdots, {T^{'}},
\end{equation}
where
$f$ is a non-linear transformation of RNN cells~(\eg LSTM~\shortcite{hochreiter1997long} or GRU~\shortcite{cho2014learning}).

Then, the probability of generating  the $t$-th token $y_t$ conditioned on the input sequence $\vec{x}$ and the previous predicted word sequence $y_{1:t-1}$
is computed by
\begin{equation}
\begin{array}{ll}
\Pr \left( y_t|y_{1:t-1}, \vec{x}\right)
=g(\vec{y}_{t-1},\vec{s_t},\vec{c_t})
,
\end{array}
\end{equation}
where $g(.)$ is a function (\eg $\mbox{softmax}$) to produce valid probability distribution for sampling the next word of the output sequence.

\subsection{Problem Definition and Overview}
In this paper, our task is to perceive the emotion involved in the input post and incorporate it into the generation process for \emph{automatically}
producing both \emph{semantically reasonable} and \emph{emotionally appropriate} response.
Hence, our conversation generation problem is defined as, given an input post $\vec{x}= \{x_1,x_2,\cdots, x_{T}\}$ with its emotion $\vec{e_p}$, for a dialogue system, generate a corresponding response sequence $\vec{y} = \{y_1,y_2,\cdots, y_{T^{'}}\}$ with proper emotion $\vec{e_r}$.

To address the problem, we propose an emotion-aware chat machine (\textsf{EACM}), which primarily consists of two subcomponents, the emotion selector and the response generator. More concretely, the emotion selector is in charge of the emotion selecting process, yielding an emotion distribution over $\mathcal{E}_{c}$ for the to-be-generated response based on the input post and its emotion:
\begin{equation}
\vec{e_r}^{*}\leftarrow  \mathop{\arg\max}_{\vec{e_r}\in \mathcal{E}_{c}} \Pr(\vec{e_r}|\vec{x},\vec{e_p}),
\end{equation}
where $\mathcal{E}_{c}$ is the vector space of the emotions. Subsequently, the generator generates a corresponding response to the given input based on
the obtained emotion $\vec{e_r}^{*}$ and the input post $\vec{x}$, namely,
\begin{equation}
\vec{y_t}^{*}\leftarrow \mathop{\arg\max}\Pr(\vec{y_{t}}|\vec{y}_{1:t-1},\vec{x},\vec{e_r}^{*}).
\end{equation}

\paratitle{Remark}.
Actually, previous approaches usually assume that the emotion of the generated response is derived from a unique emotion category\footnote{Here we follow the work~~\cite{zhou2018emotional}, where the emotion categories are~\{Angry, Disgust, Happy, Like, Sad, Other\}.}~(denoted by the one-hot vectors in $\mathcal{E}_{c}$). However, human emotions are intuitively more delicate, and thus we argue that the emotion of each response may not be limited in a single emotion category.
To this end, we assume that the emotion probability distribution is over the entire vector space $\mathcal{E}_{c}$.

\subsection{Self-attention Enhanced Emotion Selector}
\subsubsection{Self-attention Based Encoding Network}

\begin{figure}[!t]
\centering
\includegraphics[scale=0.5]{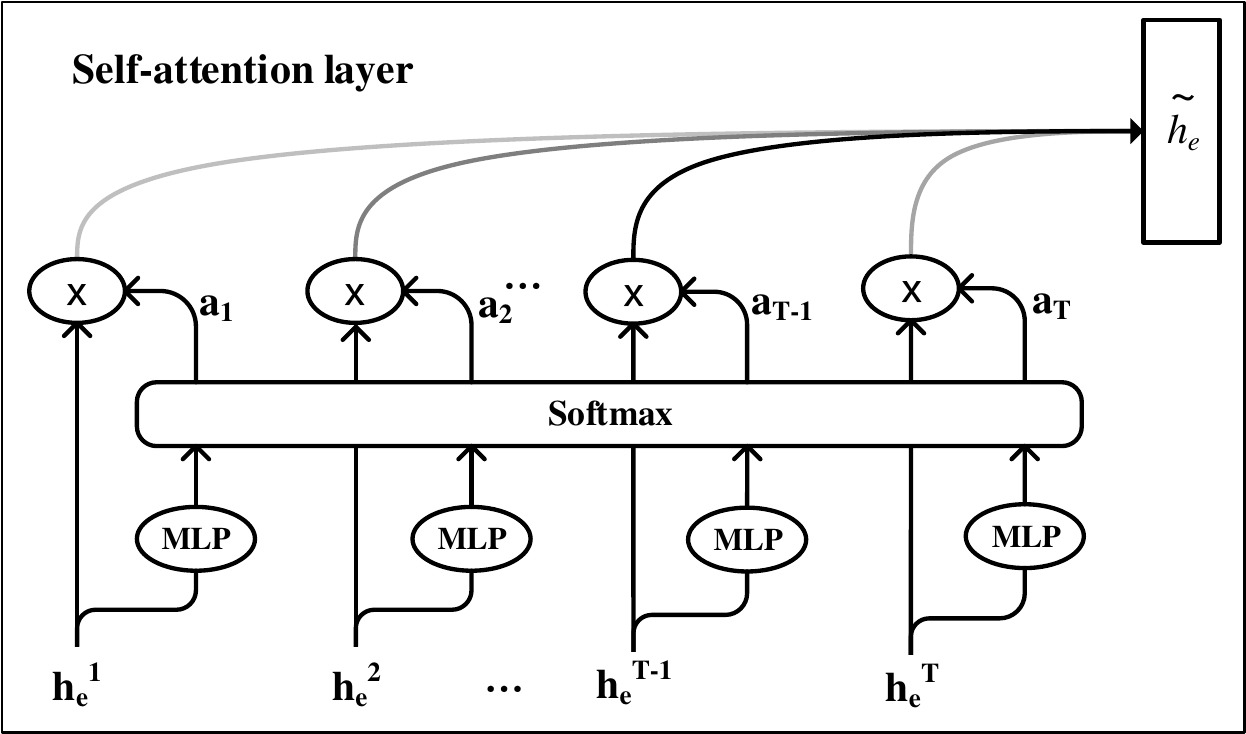}
\caption{Details of self-attention layer for emotion encoder.}
\label{fig3}
\vspace{-0.5cm}
\end{figure}

Our information encoding network consists of two parts: an emotion encoder and a semantics encoder. We leverage these two encoders by explicitly extracting emotional information and semantic information, and then feed them into a fusion prediction network.
Specifically, the emotion encoder is implemented using a \emph{GRU} network to extract information from post sequence $\vec{x}= (x_1,x_2,\cdots, x_{T})$,
and map them into hidden representations $\vec{h_e} = (\vec{h}_e^1,\vec{h}_e^2,\cdots, \vec{h}_e^{T})$ using the following formula:
\begin{equation}
\vec{h}_e^t = \mathbf{GRU}(\vec{h}_e^{t-1},x_t),
\end{equation}
where $\vec{h}_e^t$ denotes the hidden state of post's emotion information at time step $t$.

To enhance the representation power of the hidden states, we utilize \emph{self-attention} mechanism~\cite{lin2017structured} to enable the encoders to attend to emotion-rich words in the post, and then obtain the emotional hidden state $\widetilde{\vec{h}_e}$ by calculating:
\begin{equation}
\widetilde{\vec{h}_e} = \sum_{i=1}^{T}a_i\vec{h}_e^i,
\end{equation}
where $a_i$ is the weight of hidden state $\vec{h_e^i}$, which is calculated by feeding $\vec{h_e^i}$ into a multi-layer perceptron with a softmax layer to ensure that all the weights sum up to 1:
\begin{equation}
a_i = softmax(\vec{V_a}tanh(\vec{W_a}(\vec{h}_e^i)^\top).
\end{equation}
The network focuses on emotional information by imposing a cross entropy loss on the top of the emotion hidden state $\widetilde{\vec{h}_e}$, that is, passing the emotion hidden state through a linear layer and a sigmoid layer to project it into an emotion distribution over $\mathcal{E}_{c}$, and then calculating the cross entropy loss as follows,
\begin{equation}
\oldhat{\vec{e}_p} = \sigma(\vec{W_e}\widetilde{\vec{h}_e} + b),
\end{equation}
\begin{equation}
\mathcal{L}_p = - \vec{e}_p {\rm log}(\oldhat{\vec{e}_p}),
\end{equation}
where $\vec{e}_p$ is a multi-hot representation of post's emotion vector since it may contain various emotions and $\mathcal{L}_p$ is the loss function.

However, simply mapping $\oldhat{\vec{e}_p}$ to the response emotion category $\vec{e_r}$ is insufficient to model the emotion interaction process between partners,
as we cannot choose emotion category only based on post's emotion category under some circumstances. In fact, some posts expressing negative feelings like sad are inappropriate to be replied with the same emotion, such as ``It's a pity you can't come with us'' or ``I'm so sad that you broke my heart''.
Therefore, we not only consider the post's emotion information, but also take into account its semantic meaning by combing another GRU network (\ie semantics encoder) to encode post's semantic information for generation. Similarly, we get a weighted summation of the hidden states represented as $\widetilde{\vec{h}_s}$.

\subsubsection{Emotional and Semantic Word Embeddings}
In order to force the emotion selector to focus on  different aspects of auxiliary information of the given post, we apply the emotion embedding for the emotional encoder and the semantic embedding for the semantic encoder, respectively.
In particular, we make use of sentiment specific word embedding ({\it SSWE})~\cite{tang2014learning} and {\it word2vec} embedding~\cite{mikolov2013efficient} in our model. More concretely, {\it SSWE} encodes sentiment information in the continuous representation of words, mapping words with the same sentiment to the neighbor word vectors, which is used in the emotion encoder to promote the ability of perceiving emotional information in post's utterance.
Simultaneously, {\it word2vec} is used to extract semantic information from the post, and the two embeddings work interactively to guarantee the efficacy of the encoding network.

\subsubsection{Fusion and Prediction Network}
To construct the response emotion category $e_r$, we consider to use a fusion network to balance the contributions derived from different types of information, and employ a prediction network to select the response emotion categories based on such mixed information.
Then we concatenate the obtained $\widetilde{\vec{h}_s}$ and $\widetilde{\vec{h}_e}$ and feed it into a sigmoid layer to yield a trade-off weight:
\begin{align}
w &= \sigma([\widetilde{\vec{h}_s};\widetilde{\vec{h}_e}]),\\
\widetilde{\vec{h}_e^{'}}& = tanh(\widetilde{\vec{h}_e}), \\
\widetilde{\vec{h}_s^{'}}& = tanh(\widetilde{\vec{h}_s}) .
\end{align}
The final representation is a weighted sum of the semantic hidden state and the emotional hidden state:
\begin{equation}
\widetilde{\vec{h}_{es}} = w \otimes \widetilde{\vec{h}_s^{'}} + (1-w) \otimes \widetilde{\vec{h}_e^{'}},
\end{equation}
where $\otimes$ indicates element-wise multiplication.
The final representation is fed into a prediction network to produce an emotion vector for generation, which is passed through MLP and then mapped into a probability distribution over the emotion categories:
\begin{equation}
\oldhat{\vec{e}_r} = \sigma(\vec{W_r}\widetilde{\vec{h}_{es}} + b),
\end{equation}
\begin{equation}
\mathcal{L}_r = - \vec{e}_r {\rm log}(\oldhat{\vec{e}_r}),
\end{equation}
where $\vec{e}_r$ is the multi-hot representation of the response emotion vector.
$\oldhat{\vec{e}_r}$ is the final response emotion vector generated through the proposed emotion selector, which is then passed to the generator for emotional response generation.
Intuitively, the emotion selector can adaptively determine the appropriate emotion in the emotion selection process for emotional response generation, by taking into account both the post's semantic information and emotional information.

\subsection{Emotion-Biased Response Generator}
To construct the generator, we consider to use an emotion-enhanced seq2seq model that is capable of balancing the emotional part with the semantic part and generate intelligible responses.

Thereby, we first generate the response emotion embedding $\vec{V}_e$ by multiplying $\oldhat{\vec{e}_r}$ with a randomly initialized matrix:
\begin{equation}
\vec{V}_e = \matrix{W}_e\oldhat{\vec{e}_r},
\end{equation}
where $\matrix{W}_e$ is the emotion embedding matrix, which is the latent emotional factors, \ie the high-level abstraction of emotion expressions by following Plutchik's assumptions~\shortcite{plutchik1980general}.
As mentioned, a one-hot emotion embedding is inappropriate and thus here we do not use a softmax on $\oldhat{\vec{e}_r}$ to only pick an optimal emotion category
for generation.
As such, we call it as \emph{soft-emotion injection} procedure, which is used to model the diversity of emotions.

By following the work \cite{vinyals2015grammar}, we use a new encoder to encode $\vec{x}$ for obtaining a sequence of hidden states $\vec{h} = (\vec{h}_1,\vec{h}_2,\cdots, \vec{h}_T)$ through a RNN network,
and then generate the context vector $\vec{c}_t$ for decoding the current hidden state $\vec{s_t}$, via applying attention mechanism to re-assign an attended weight to each encoder hidden state $\vec{h}_i$.
\begin{eqnarray}
u^t_i &=& \vec{v^\top} \tanh (\vec{W}_1 \vec{h}_i + \vec{W}_2 \vec{s}_t), \label{eqn:attn1} \\
a^t_i &=& \mathrm{softmax}(u^t_i) \label{eqn:attn2},\\
\vec{c}_t &=& \sum_{i=1}^{T} a^t_i  \vec{h}_i.
\end{eqnarray}

At each time step $t$, the context vector encoded with attention mechanism enable our model to proactively search for salient information which is important for decoding over a long sentence.
However, it neglects the emotion ($\vec{V}_e$) derived from the response during generation, and thus we propose an emotion-biased attention mechanism
to rewritten Eq.(\ref{eqn:attn1}),
\begin{equation}
u^t_i = \vec{v^\top} \tanh (\vec{W}_1 \vec{h}_i + \vec{W}_2 \vec{s}_t + \vec{W}_3 \vec{V}_e).
\end{equation}
The context vector $\vec{c}_t$ is concatenated with $\vec{s}_t$ and forms a new hidden state $\vec{s}_t^{'}$:
\begin{equation}
\vec{s}_t^{'} = \vec{W}_{4}[\vec{s}_t ; \vec{c}_t],
\end{equation}
from which we make the prediction for each word;
$\vec{s}_t$ is obtained by changing Eq.~\eqref{eq:over_opt} into:
\begin{equation}
\vec{s}_t = \mathbf{GRU}(\vec{s}_{t-1}^{'}, [ \vec{y}_{t-1} ; \vec{V}_e ]) ,
\end{equation}
which fulfills the task of injecting emotion information while generating responses. To be consistent with previous conversation generation approaches,
here we consider to use \emph{cross entropy} to be the loss function, which is defined by
\begin{align}
\mathcal{L}_{seq2seq}(\theta)&= -logP(\vec{y|x})  \\
&= -\sum_{t=1}^{T^{'}} log P(y_t \mid y_1, y_2,\cdots, y_{t-1}, \vec{c}_t,\vec{V}_e ) ,
\end{align}
where $\theta$ denotes the parameters.

\subsection{Loss Function}

The loss function of our model is a weighted summation of the semantic loss and sentiment loss:
\begin{equation}
\mathcal{L}_{EACM}(\theta) = \alpha \mathcal{L}_e + (1-\alpha)\mathcal{L}_{seq2seq},
\end{equation}
where $\alpha$ is a balance factor, and $\mathcal{L}_e$ denotes the \emph{emotional} loss, namely,
\begin{equation}
\mathcal{L}_e = \mathcal{L}_p + \mathcal{L}_r.
\end{equation}

\section{Experimental Results}
\label{sec:exp}

\subsection{Dataset}

\renewcommand\arraystretch{1.4}
\begin{table}[!t]
\small
\begin{tabular}{|c|c|c|c|}
\hline
\multirow{4}{*}{\textbf{Training $\#$}} & \multicolumn{2}{c|}{\textbf{Posts}}                     & 219,162   \\ \cline{2-4}
                          & \multirow{3}{*}{\textbf{Responses}} & \textbf{\textit{No Emotion}} & 1,586,065 \\ \cline{3-4}
                          &                            & \textbf{\textit{Single Emotion}  } & 2,792,339 \\ \cline{3-4}
                          &                            & \textbf{\textit{Dual Emotion} }& 53,545    \\ \hline
\textbf{Validation $\#$}                & \multicolumn{3}{c|}{1,000}                                 \\ \hline
\textbf{Testing $\#$}                   & \multicolumn{3}{c|}{1,000}                                 \\ \hline
\end{tabular}
\caption{Details of ESTC dataset. }
\label{tab:1}
\vspace{-0.5cm}
\end{table}

\begin{figure}[!t]
\centering
\includegraphics[scale=0.4]{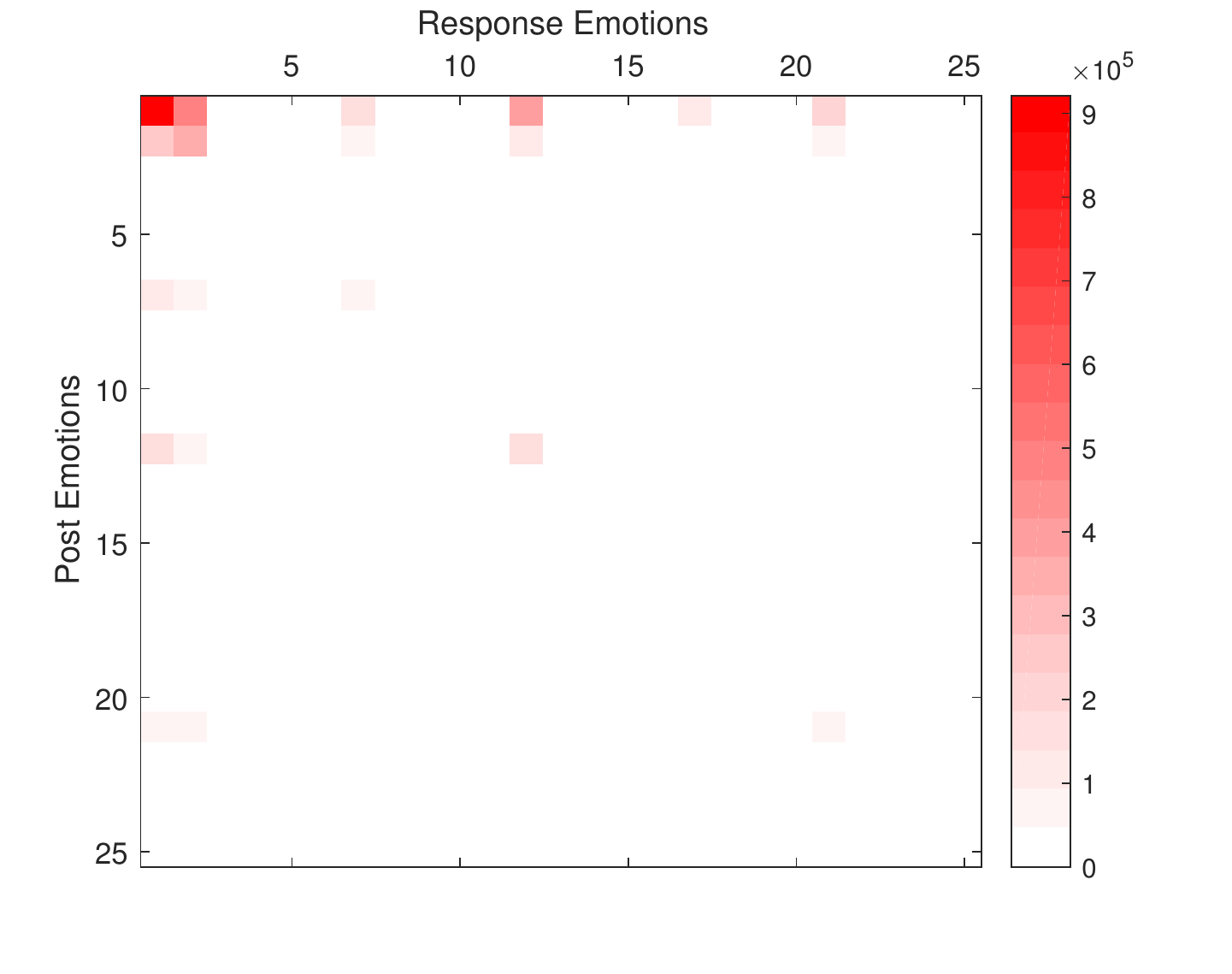}
\vspace{-0.35cm}\caption{An illustration of emotion interaction pattens (EIPs). Numerical values on each axis represent different emotion categories.  The darker each grid is, the more pairs ($\vec{e_p}$,$\vec{e_r}$) appear in such grid. }
\label{fig3}
\vspace{-0.5cm}
\end{figure}
We conduct our experiments on a public dataset, \ie Emotional Short-Text Conversation (ESTC) derived from STC \shortcite{shang2015neural}, to evaluate our experimental results. In particular, we follow the work~\cite{zhou2018emotional} to train an emotion classifier for assigning emotional labels to the sentences in the dataset.

\paratitle{ESTC}, which contains over four million real-world conversations obtained from Chinese micro-blogging, \ie Weibo.
The raw dataset contains $4,433,949$ post-comment pairs, from which $1,000$ pairs are extracted for validation and another $1,000$ pairs are used for testing. Details of this dataset is illustrated in Table \ref{tab:1}.

\paratitle{Preprocessing}.
As the raw dataset (STC) does not have emotion labels, so we train an emotion classifier based on  BERT \cite{devlin2018bert} model over two different datasets,
\ie NLPCC 2013\footnote{\url{http://tcci.ccf.org.cn/conference/2013/}} and NLPCC 2014\footnote{\url{http://tcci.ccf.org.cn/conference/2014/}} emotion classification datasets by following~\cite{zhou2018emotional}, which contain $29,417$ manually annotated data in total,
and the best performance(accuracy) of $0.7257$ is achieved at the $19,635$ step.
Specifically, each sentence is marked with two labels, namely, a primary label and a secondary one.
We preprocess the labels over the mentioned emotion categories, \ie (\emph{like, disgust, sad, angry, happy, other}), note that here ``\emph{other}'' indicates no any emotion information, and rare emotion categories like {\it fear} are removed.
In particular, unlike ~\cite{zhou2018emotional} using solely one label for classification, we consider both of the emotion labels and thus regard it as a multi-label classification task.
Under such circumstance, there are three cases appearing in the label sets, \ie no emotion labeled with (other, other), one emotion labeled with (emo1, other), as well as two emotions labeled with (emo1, emo2), respectively.

To evaluate the emotion perception ability of different approaches over the emotion categories,
we build an emotion-rich dialogue set for a fair empirical comparison.
Specifically, we randomly chose $1,000$ pairs whose primary emotion label is among the (\emph{like, disgust, sad, angry or happy}) categories,
with $200$ pairs for each emotion, respectively.
In addition, we also present an in-depth analysis the \emph{emotion interaction pattern} (EIP) over conversations, in which each EIP is defined as a ($\vec{e_p}$,$\vec{e_r}$) pair for each conversation to reflect the transition pattern from the post's emotion to the response's emotion.
Figure \ref{fig3} shows a heatmap to depict the number of EIPs appearing in the dataset, and each row (\emph{or} column) indicates the post's emotion $\vec{e_p}$ (\emph{or} the response's emotion $\vec{e_r}$).
From the figure we can observe that:~(1) The darker each grid is, the more pairs ($\vec{e_p}$,$\vec{e_r}$) appearing in such grid;
(2) The heat map is sparse since the EIPs of some post-response emotion pairs are overly rare to appear in our dataset.

Moreover, we also leverage ``Weibo Sentiment Dataset'' provided by Shujutang\footnote{\url{http://www.datatang.com/data/45439}} to train the sentiment-specific word embeddings (SSWE), which consists of two million Weibo sentences with sentiment labels, and we remove some extra domain-specific punctuations like ``@user'' and ``URLs''.

\subsection{Evaluation Metric}
As reported in~\cite{liu2016not}, BLEU might be improper to evaluate the conversation generation problem, as it correlates weakly with human judgements of the response quality, similar situations for METEOR~\cite{banerjee2005meteor} and ROUGE~\cite{lin2004rouge}.
Besides, there still exists a challenge of automatically evaluating the generation model from emotion perspective.
As a result, in this paper we adopt {\it distinct-1} and {\it distinct-2} by follow the work~\cite{li2015diversity} to be as the metrics for
evaluating the diversity of the generated responses, which measures the degree of diversity by computing the number of distinct uni-grams and bi-grams
in the generated responses, and can indirectly reflect the degree of emotion diversity, as the generated sentence containing diverse emotions is more likely to have more abundant words in principle.
In addition, we also carry out a manual evaluation for evaluate the performance of the generated responses at \emph{emotional}-level and \emph{semantic}-level separately with human intuition, and then the {\it response quality} is calculated by combining such two results at different levels for integrally assessing different models.

\paratitle{Automatic Evaluation}.
As mentioned, we consider to use {\it distinct-1} and {\it distinct-2}~\cite{li2015diversity} to be as our automatic evaluation metric.  {\it Distinct-n} is defined as the number of distinct n-grams in generated responses. The value is scaled by total number of generated tokens.

\paratitle{Human Evaluation}.
We randomly sampled $200$ posts from the \emph{test} set, and then aggregate the corresponding responses returned by each evaluated method, then three graduate students~(whose research areas are not in text processing area) are invited for labeling.
Each generated response is labeled from two different aspects, \ie \emph{emotion} and \emph{semantics}.
Specifically, from emotion perspective, each generated response is labeled with (score 0) if its emotion is apparently inappropriate (namely evident emotion collision,\eg angry-happy) to the given post, and (score 1) otherwise. From semantic perspective, we evaluate the generated results using the scoring metrics as follows. Note that if conflicts happen, the third annotator determines the final result.
\begin{itemize}
    \item 1: If the generated sentence can be obversely considered as a appropriate response to the input post;
    \item 0: If the generated sentence is hard-to-perceive or has little relevance to the given post.
\end{itemize}

To conduct an integral assessment of the models at both \emph{emotional}-level and \emph{semantic}-level, we measure the {\it response quality} by using the formula as follows,
\begin{equation}
Q_{response} = S_{sentiment} \land S_{semantics},
\end{equation}
where $Q_{response}$ reflects the {\it response quality} and $S_{sentiment}$, $S_{semantics}$ denote the sentiment score and semantic score, respectively, which is used to  means the {\it response quality} of each case is equal to 1 if and only if both of its sentiment score and semantic score are scored as 1.

\subsection{Baselines}

We compare our model with the following baselines.

\paratitle{Seq2seq}~\cite{sutskever2014sequence}, the traditional {\it Seq2seq} model is adopted as one of our baselines.

\paratitle{ECM}~\cite{zhou2018emotional}, as mentioned, {\it ECM} model is improper to directly be as the baseline since it cannot automatically select an appropriate emotion label to the respond. Thereby, we manually designate a most frequent response emotion to {\it ECM} for fairness comparison. Specifically, we train a post emotion classifier to automatically detect post's emotion, and then choose the corresponding response emotion category using the most frequent response's emotion to the detected post's emotion over EIPs.

\paratitle{Seq2seq-emb}~\cite{zhou2018emotional,huang2018automatic}, \emph{Seq2seq} with emotion embedding ({\it Seq2seq-emb}) is also adopted in the same manner. This model encode the emotion category into an embedding vector, and then utilize it as an extra emotion input when decoding.

Intuitively, the generated responses from {\it ECM} and {\it Seq2seq-emb} can be viewed as the indication of the performance of simply incorporating the \emph{EIP}s
for modeling the emotional interactions among the conversation pairs.

\subsection{Implementation Details}
For all approaches, each encoder and decoder with $2$-layers GRU cells containing 256 hidden units, and all of the parameters are not shared between such two different layers. The vocabulary size is set as $40,000$, and the OOV (out-of-vocabulary) words are replaced with a special token {\it UNK}. The size of word embeddings is $200$, which are randomly initialized. The emotion embedding is a $6\times 200$-dimensional matrix (if used). The parameters of {\it imemory} and {\it ememory} in {\it ECM} are the same as the settings in  \cite{zhou2018emotional}.
We use stochastic gradient descent (SGD) with mini-batch for optimization when training, and the batch size and the learning rate are set as $128$ and $0.5$, respectively. The greedy search algorithm is adopted for each approach to generate responses.
Additionally, for speeding up the training process, we leverage the well-trained {\it Seq2seq} model to initialize other methods.

The parameters for our proposed method are empirically set as follows:
\emph{SSWE} is trained by following the parameter settings in~\cite{tang2014learning}.
The length of hidden layer is set at $20$, and we use AdaGrad \cite{Duchi2011Adaptive} to update the trainable parameters and the learning rate is set as $0.1$.
The size of emotion embedding and word embedding are both set at $200$.
In particular, the \emph{Word2vec} embedding is used based on {\it Tencent AI Lab Embedding\footnote{\url{https://ai.tencent.com/ailab/nlp/embedding.html}}}, which is pre-trained over $8$ million high-quality Chinese words and phrases by using directional skip-gram method \cite{song2018directional}.
We use {\tt jieba\footnote{\url{https://github.com/fxsjy/jieba}}} for word segmentation during the evaluation process.

\subsection{Results and Discussion}
\label{subsec:results}
In this section, we evaluate the effectiveness of generating
emotional responses by our approach as comparison to the
baseline methods.

\paratitle{Automatic Evaluation}.
From Table \ref{tab2}, we can observe that:
(i) \emph{ECM} performs worse than \emph{Seq2seq}, the reason might be the emotion selection process is based on two-stage process, \ie post emotion detection process
and response emotion selection process, which would significantly reduce the diversity and quality of emotion response generation due to the errors of emotion classification and the transition pattern modeling procedure.
In particular, the emotion category {\it (other, other)} is more likely to be chosen than other emotion categories. We will present an in-depth analysis in Section~\ref{sec:exp-case-study}.
and (ii) Our proposed \emph{EACM} consistently outperforms all of the baselines in terms of {\it distinct-1} and {\it distinct-2}. The results demonstrate that our emotion selection process is really effective in enhancing the ability of generating more diverse words.

\begin{table}[!htp]
\centering
\begin{tabular}{lcc}
\hline
\textbf{Models}      & \multicolumn{1}{l}{\textbf{Distinct-1}} & \multicolumn{1}{l}{\textbf{Distinct-2}} \\ \hline
\textbf{Seq2seq}     & 0.0608                          & 0.2104                             \\
\textbf{Seq2seq-emb} & 0.0628                          & 0.2370                                 \\
\textbf{ECM}         &0.0551               & 0.2022                                \\
\textbf{EACM}        & \textbf{0.0745 }                         & \textbf{0.2749}                    \\ \hline
\end{tabular}
\caption{{\bf Automatic evaluation:} distinct-1 and distinct-2.}
\vspace{-0.5cm}
\label{tab2}
\end{table}

\paratitle{Human Evaluation}.
From Table \ref{tab3}, we can observe that:
\emph{Seq2seq-emb} provides the worst performance as expected, this is because the generation process apparently interrupted would significantly reduce the accuracy and quality of generating responses and thus generate some hard-to-perceive sentences.
\emph{ECM} achieves a relatively better result, as it is good at modeling the emotion dynamics when decoding (\ie \emph{internal memory})
and assigning different generation probabilities to emotion/generic words for explicitly modeling emotion expressions (\ie \emph{external memory}).
Specifically, \emph{ECM} results in a remarkable improvement over \emph{Seq2seq} in terms of \emph{sentiment score}, but it performs poorly when comparing \emph{semantic score}, which demonstrates the effectiveness of \emph{emotion injection}, however the explicit two-stage procedure might reduce the smoothness of generated responses with low \emph{semantic score}.
Our proposed model \emph{EACM}  consistently outperforms all baseline methods, and the improvements are statistically significant on all metrics.
For example, \emph{EACM} outperforms \emph{ECM} by $16.9\%$, $1.72\%$ and $25.81\%$ in terms of  \emph{semantic score},
\emph{sentiment score} and \emph{response quality}, respectively.
The reason might due to the fact that \emph{EACM} is capable of simultaneously encoding the
semantics and the emotions in a post for generating appropriately expressed emotional response within a unified end-to-end neural architecture,
which are benefit for alleviating the \emph{hard emotion injection} problem~(as compared to \emph{soft emotion injection}).

The percentage of the sentiment-semantics scores under the human evaluation is shown in Table \ref{tab4}.
For \emph{ECM}, the percentage of (0-0) degrades while the percentage of (1-0) increases as opposed to \emph{Seq2seq}, which suggests that the effectivenss of EIP, \ie the most frequent response emotions have low probability to result in emotional conflicts, In addition, the percentage of (1-1) degrades while the percentage of (1-0) increases, which reflects that directly using emotion classifier to model emotion interaction process is insufficient.
In comparison, \emph{EACM} reduces the percentage of generating responses with wrong emotion and correct semantics (\ie 0-1)
while increase the percentage of (1-1) correspondingly, which demonstrate that \emph{EACM} is capable of successfully model the emotion interaction pattern among human conversations and meanwhile guarantee semantic correctness.

\renewcommand\arraystretch{1.2}
\begin{table}[]
\centering
\begin{tabular}{lccc}
\hline
\textbf{Models}      &\textbf{$\bar{\bm{S}}_{\text{semantics}}$} &\textbf{$\bar{\bm{S}}_{\text{sentiment}}$}
 & \textbf{$\bar{\bm{Q}}_{\text{response}}$}  \\ \hline
\textbf{Seq2seq}     & 0.390          & 0.815           & 0.360        \\
\textbf{Seq2seq-emb} &    0.280       &  0.795         & 0.250         \\
\textbf{ECM}         & 0.355          & 0.870  & 0.310         \\
\textbf{EACM}        & \textbf{0.415} & \textbf{0.885}  & \textbf{0.390}      \\ \hline
\end{tabular}
\caption{Human evaluation: averaged semantic score, sentiment score and response quality.}
\vspace{-0.45cm}
\label{tab3}
\end{table}

\begin{table}[!t]
\begin{tabular}{lcccc}
\hline
\textbf{Method(\%)}  & 1-1         & 1-0           & 0-1          & 0-0          \\ \hline
\textbf{Seq2seq}     & 36          & 45.5          & 3            & 15.5         \\
\textbf{Seq2seq-emb} & 25          &   54.5        &  3           & 17.5          \\
\textbf{ECM}         & 31          & \textbf{56}            & 4.5          & \textbf{8.5} \\
\textbf{EACM }       & \textbf{39} & 49.5 & \textbf{2.5} & 9            \\ \hline
\end{tabular}
\caption{The percentage of the sentiment-semantic score given by human evaluation.}
\vspace{-0.5cm}
\label{tab4}
\end{table}

\begin{figure*}[htbp]
\centering
\includegraphics[scale=0.54]{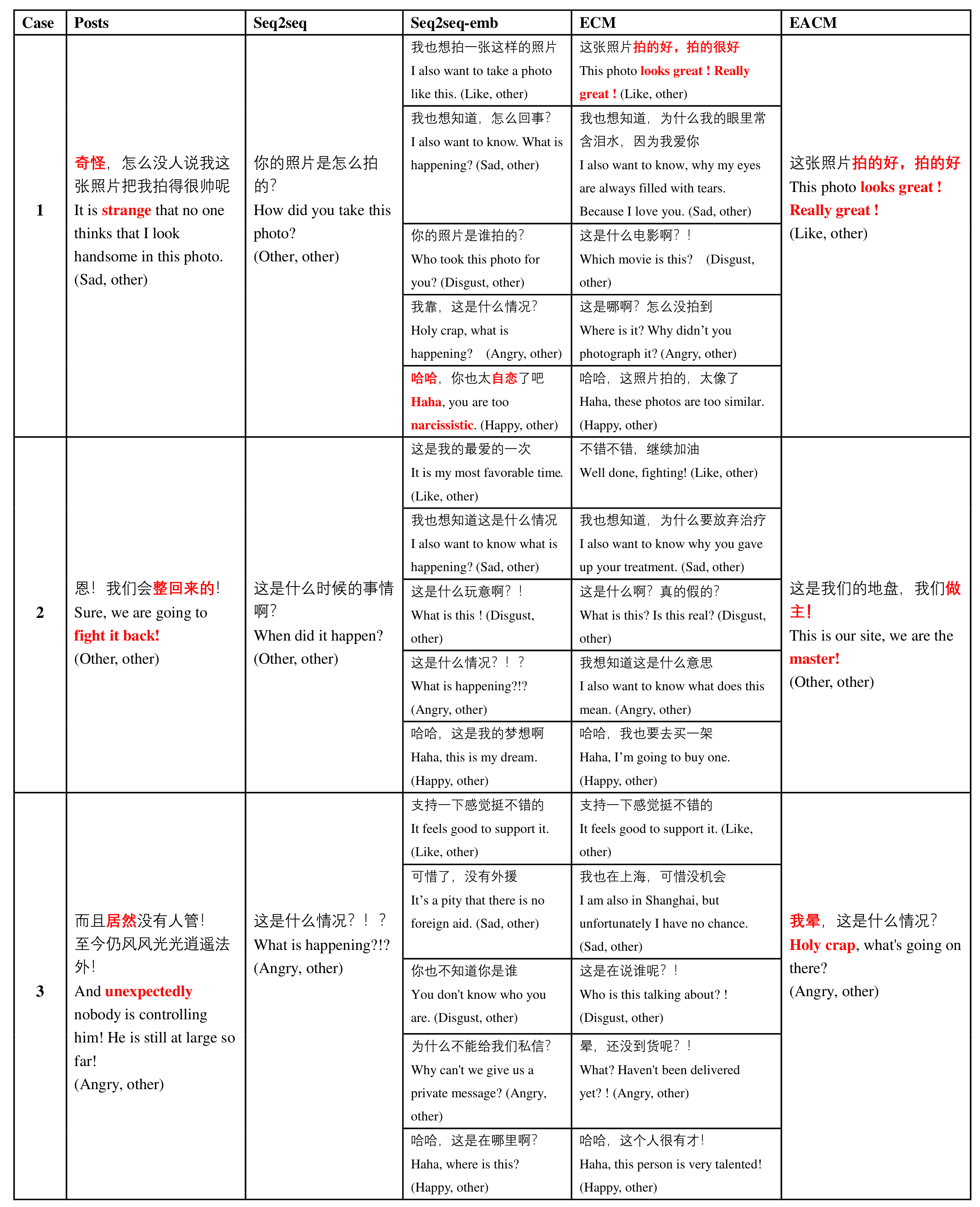}
\caption{Case Study. $3$ Samples (with the given posts and the corresponding responses) generated by \emph{Seq2seq}, \emph{Seq2seq-emb}, \emph{ECM} and \emph{EACM}.
Words that express appropriate emotion in responses are highlighted in red,  along with their posts' corresponding emotion words.}
\label{fig5}
\vspace{-0.5cm}
\end{figure*}

\subsection{Case Study}
\label{sec:exp-case-study}
In this section, we present an in-depth analysis of emotion-aware response generation results of our proposed approach. We select $3$ samples (with the input posts and their corresponding responses) generated by different methods are shown in Figure~\ref{fig5}, as can be seen:

\paratitle{Case 1}.
The results of \emph{EACM} and \emph{ECM} with (like, other) are similar and correct. However, responses given by \emph{ECM} with other emotions are improper in semantics. Similar scenario for \emph{Seq2seq-emb} with (Happy, other). However, the most frequent emotion to (Sad, other) is never be (Happy, other) over EIPs, and thus EIPs would fail  to achieve the task.

\paratitle{Case 2}.
\emph{EACM} can generate the relevant emotional response \ie (Other, other) for the post with (Other, other).
However, all of \emph{ECM} with different emotions seems improper for response (especially for \emph{ECM} with (Happy,other)), which reflects directly using a designated emotion for generation might be a unreasonable way for modeling the emotion interaction pattern. In addition, \emph{Seq2seq} cannot detect the emotion and thus generate a irrelevant response. As most of conversations belong to (other, other), and thus the diversity of such emotion category is quite  complicated and hard for training.

\paratitle{Case 3}. The emotion in the case is (Angry, other), however the responses provided by \emph{ECM} with (Angry, other) is obviously incorrect in semantics, which demonstrate that simply using post's emotion is inappropriate for response generation. As compared,  \emph{EACM} generates the correct emotional response with a emotion-specific word (\ie ``Holy crap''), which demonstrates \emph{EACM} is good at controlling sentence emotion for generation.

\section{Conclusion}
\label{sec:con}
In this paper, we propose an emotion-aware chat machine (EACM) to address the emotional response generation problem, which is composed of an emotion selector and a response generator. Specifically, a unified fusion-prediction network with self-attention mechanism is also developed to supervise the emotion selection process for generating a emotion-biased response. Extensive experiments conducted on a public dataset demonstrate the effectiveness of our proposed method as compared to baselines
at both \emph{semantic}-level and \emph{emotional}-level, in terms of automatic evaluation and human evaluation metrics.

\begin{acks}
  This work was supported in part by the National Natural Science Foundation of China under Grant No.61602197, and in part by  Equipment Pre-Research Fund for The 13th Five-year Plan under Grant No.41412050801.
\end{acks}

\bibliographystyle{ACM-Reference-Format}
\balance
\bibliography{sample-sigconf}


\begin{thebibliography}{48}


\ifx \showCODEN    \undefined \def \showCODEN     #1{\unskip}     \fi
\ifx \showDOI      \undefined \def \showDOI       #1{#1}\fi
\ifx \showISBNx    \undefined \def \showISBNx     #1{\unskip}     \fi
\ifx \showISBNxiii \undefined \def \showISBNxiii  #1{\unskip}     \fi
\ifx \showISSN     \undefined \def \showISSN      #1{\unskip}     \fi
\ifx \showLCCN     \undefined \def \showLCCN      #1{\unskip}     \fi
\ifx \shownote     \undefined \def \shownote      #1{#1}          \fi
\ifx \showarticletitle \undefined \def \showarticletitle #1{#1}   \fi
\ifx \showURL      \undefined \def \showURL       {\relax}        \fi
\providecommand\bibfield[2]{#2}
\providecommand\bibinfo[2]{#2}
\providecommand\natexlab[1]{#1}
\providecommand\showeprint[2][]{arXiv:#2}

\bibitem[\protect\citeauthoryear{Anderson, He, Buehler, Teney, Johnson, Gould,
  and Zhang}{Anderson et~al\mbox{.}}{2018}]%
        {anderson2018bottom}
\bibfield{author}{\bibinfo{person}{Peter Anderson}, \bibinfo{person}{Xiaodong
  He}, \bibinfo{person}{Chris Buehler}, \bibinfo{person}{Damien Teney},
  \bibinfo{person}{Mark Johnson}, \bibinfo{person}{Stephen Gould}, {and}
  \bibinfo{person}{Lei Zhang}.} \bibinfo{year}{2018}\natexlab{}.
\newblock \showarticletitle{Bottom-up and top-down attention for image
  captioning and visual question answering}. In
  \bibinfo{booktitle}{\emph{CVPR}}. \bibinfo{pages}{6077--6086}.
\newblock


\bibitem[\protect\citeauthoryear{Bahdanau, Cho, and Bengio}{Bahdanau
  et~al\mbox{.}}{2015}]%
        {bahdanau2014neural}
\bibfield{author}{\bibinfo{person}{Dzmitry Bahdanau},
  \bibinfo{person}{Kyunghyun Cho}, {and} \bibinfo{person}{Yoshua Bengio}.}
  \bibinfo{year}{2015}\natexlab{}.
\newblock \showarticletitle{Neural machine translation by jointly learning to
  align and translate}. In \bibinfo{booktitle}{\emph{ICLR}}.
\newblock


\bibitem[\protect\citeauthoryear{Banerjee and Lavie}{Banerjee and
  Lavie}{2005}]%
        {banerjee2005meteor}
\bibfield{author}{\bibinfo{person}{Satanjeev Banerjee} {and}
  \bibinfo{person}{Alon Lavie}.} \bibinfo{year}{2005}\natexlab{}.
\newblock \showarticletitle{METEOR: An automatic metric for MT evaluation with
  improved correlation with human judgments}. In \bibinfo{booktitle}{\emph{ACL
  workshop}}. \bibinfo{pages}{65--72}.
\newblock


\bibitem[\protect\citeauthoryear{Callejas, Griol, and
  L{\'o}pez-C{\'o}zar}{Callejas et~al\mbox{.}}{2011}]%
        {callejas2011predicting}
\bibfield{author}{\bibinfo{person}{Zoraida Callejas}, \bibinfo{person}{David
  Griol}, {and} \bibinfo{person}{Ram{\'o}n L{\'o}pez-C{\'o}zar}.}
  \bibinfo{year}{2011}\natexlab{}.
\newblock \showarticletitle{Predicting user mental states in spoken dialogue
  systems}.
\newblock \bibinfo{journal}{\emph{EJASP}} (\bibinfo{year}{2011}),
  \bibinfo{pages}{6}.
\newblock


\bibitem[\protect\citeauthoryear{Cho, Van~Merri{\"e}nboer, Gulcehre, Bahdanau,
  Bougares, Schwenk, and Bengio}{Cho et~al\mbox{.}}{2014}]%
        {cho2014learning}
\bibfield{author}{\bibinfo{person}{Kyunghyun Cho}, \bibinfo{person}{Bart
  Van~Merri{\"e}nboer}, \bibinfo{person}{Caglar Gulcehre},
  \bibinfo{person}{Dzmitry Bahdanau}, \bibinfo{person}{Fethi Bougares},
  \bibinfo{person}{Holger Schwenk}, {and} \bibinfo{person}{Yoshua Bengio}.}
  \bibinfo{year}{2014}\natexlab{}.
\newblock \showarticletitle{Learning phrase representations using RNN
  encoder-decoder for statistical machine translation}. In
  \bibinfo{booktitle}{\emph{EMNLP}}.
\newblock


\bibitem[\protect\citeauthoryear{Devlin, Chang, Lee, and Toutanova}{Devlin
  et~al\mbox{.}}{2018}]%
        {devlin2018bert}
\bibfield{author}{\bibinfo{person}{Jacob Devlin}, \bibinfo{person}{Ming-Wei
  Chang}, \bibinfo{person}{Kenton Lee}, {and} \bibinfo{person}{Kristina
  Toutanova}.} \bibinfo{year}{2018}\natexlab{}.
\newblock \showarticletitle{Bert: Pre-training of deep bidirectional
  transformers for language understanding}.
\newblock \bibinfo{journal}{\emph{arXiv}} (\bibinfo{year}{2018}).
\newblock


\bibitem[\protect\citeauthoryear{Duchi, Hazan, and Singer}{Duchi
  et~al\mbox{.}}{2011}]%
        {Duchi2011Adaptive}
\bibfield{author}{\bibinfo{person}{John Duchi}, \bibinfo{person}{Elad Hazan},
  {and} \bibinfo{person}{Yoram Singer}.} \bibinfo{year}{2011}\natexlab{}.
\newblock \showarticletitle{Adaptive Subgradient Methods for Online Learning
  and Stochastic Optimization}.
\newblock \bibinfo{journal}{\emph{JMLR}} \bibinfo{volume}{12},
  \bibinfo{number}{7} (\bibinfo{year}{2011}), \bibinfo{pages}{257--269}.
\newblock


\bibitem[\protect\citeauthoryear{Ghosh, Chollet, Laksana, Morency, and
  Scherer}{Ghosh et~al\mbox{.}}{2017}]%
        {ghosh2017affect}
\bibfield{author}{\bibinfo{person}{Sayan Ghosh}, \bibinfo{person}{Mathieu
  Chollet}, \bibinfo{person}{Eugene Laksana}, \bibinfo{person}{Louis-Philippe
  Morency}, {and} \bibinfo{person}{Stefan Scherer}.}
  \bibinfo{year}{2017}\natexlab{}.
\newblock \showarticletitle{Affect-lm: A neural language model for customizable
  affective text generation}.
\newblock \bibinfo{journal}{\emph{arXiv}} (\bibinfo{year}{2017}).
\newblock


\bibitem[\protect\citeauthoryear{Hochreiter and Schmidhuber}{Hochreiter and
  Schmidhuber}{1997}]%
        {hochreiter1997long}
\bibfield{author}{\bibinfo{person}{Sepp Hochreiter} {and}
  \bibinfo{person}{J{\"u}rgen Schmidhuber}.} \bibinfo{year}{1997}\natexlab{}.
\newblock \showarticletitle{Long short-term memory}.
\newblock \bibinfo{journal}{\emph{Neural computation}} \bibinfo{volume}{9},
  \bibinfo{number}{8} (\bibinfo{year}{1997}), \bibinfo{pages}{1735--1780}.
\newblock


\bibitem[\protect\citeauthoryear{Hu, Yang, Liang, Salakhutdinov, and Xing}{Hu
  et~al\mbox{.}}{2017}]%
        {hu2017toward}
\bibfield{author}{\bibinfo{person}{Zhiting Hu}, \bibinfo{person}{Zichao Yang},
  \bibinfo{person}{Xiaodan Liang}, \bibinfo{person}{Ruslan Salakhutdinov},
  {and} \bibinfo{person}{Eric~P Xing}.} \bibinfo{year}{2017}\natexlab{}.
\newblock \showarticletitle{Toward controlled generation of text}. In
  \bibinfo{booktitle}{\emph{ICML}}.
\newblock


\bibitem[\protect\citeauthoryear{Huang, Zaiane, Trabelsi, and Dziri}{Huang
  et~al\mbox{.}}{2018}]%
        {huang2018automatic}
\bibfield{author}{\bibinfo{person}{Chenyang Huang}, \bibinfo{person}{Osmar
  Zaiane}, \bibinfo{person}{Amine Trabelsi}, {and} \bibinfo{person}{Nouha
  Dziri}.} \bibinfo{year}{2018}\natexlab{}.
\newblock \showarticletitle{Automatic Dialogue Generation with Expressed
  Emotions}. In \bibinfo{booktitle}{\emph{NAACL}}. \bibinfo{pages}{49--54}.
\newblock


\bibitem[\protect\citeauthoryear{Jean, Cho, Memisevic, and Bengio}{Jean
  et~al\mbox{.}}{2014}]%
        {jean2014using}
\bibfield{author}{\bibinfo{person}{S{\'e}bastien Jean},
  \bibinfo{person}{Kyunghyun Cho}, \bibinfo{person}{Roland Memisevic}, {and}
  \bibinfo{person}{Yoshua Bengio}.} \bibinfo{year}{2014}\natexlab{}.
\newblock \showarticletitle{On using very large target vocabulary for neural
  machine translation}.
\newblock \bibinfo{journal}{\emph{arXiv}} (\bibinfo{year}{2014}).
\newblock


\bibitem[\protect\citeauthoryear{Ji, Lu, and Li}{Ji et~al\mbox{.}}{2014}]%
        {ji2014information}
\bibfield{author}{\bibinfo{person}{Zongcheng Ji}, \bibinfo{person}{Zhengdong
  Lu}, {and} \bibinfo{person}{Hang Li}.} \bibinfo{year}{2014}\natexlab{}.
\newblock \showarticletitle{An information retrieval approach to short text
  conversation}.
\newblock \bibinfo{journal}{\emph{arXiv}} (\bibinfo{year}{2014}).
\newblock


\bibitem[\protect\citeauthoryear{Li, Galley, Brockett, Gao, and Dolan}{Li
  et~al\mbox{.}}{2015}]%
        {li2015diversity}
\bibfield{author}{\bibinfo{person}{Jiwei Li}, \bibinfo{person}{Michel Galley},
  \bibinfo{person}{Chris Brockett}, \bibinfo{person}{Jianfeng Gao}, {and}
  \bibinfo{person}{Bill Dolan}.} \bibinfo{year}{2015}\natexlab{}.
\newblock \showarticletitle{A diversity-promoting objective function for neural
  conversation models}. In \bibinfo{booktitle}{\emph{NAACL}}.
\newblock


\bibitem[\protect\citeauthoryear{Li, Galley, Brockett, Spithourakis, Gao, and
  Dolan}{Li et~al\mbox{.}}{2016a}]%
        {li2016persona}
\bibfield{author}{\bibinfo{person}{Jiwei Li}, \bibinfo{person}{Michel Galley},
  \bibinfo{person}{Chris Brockett}, \bibinfo{person}{Georgios~P Spithourakis},
  \bibinfo{person}{Jianfeng Gao}, {and} \bibinfo{person}{Bill Dolan}.}
  \bibinfo{year}{2016}\natexlab{a}.
\newblock \showarticletitle{A persona-based neural conversation model}. In
  \bibinfo{booktitle}{\emph{ACL}}.
\newblock


\bibitem[\protect\citeauthoryear{Li, Monroe, and Jurafsky}{Li
  et~al\mbox{.}}{2016b}]%
        {li2016simple}
\bibfield{author}{\bibinfo{person}{Jiwei Li}, \bibinfo{person}{Will Monroe},
  {and} \bibinfo{person}{Dan Jurafsky}.} \bibinfo{year}{2016}\natexlab{b}.
\newblock \showarticletitle{A simple, fast diverse decoding algorithm for
  neural generation}.
\newblock \bibinfo{journal}{\emph{arXiv}} (\bibinfo{year}{2016}).
\newblock


\bibitem[\protect\citeauthoryear{Lin}{Lin}{2004}]%
        {lin2004rouge}
\bibfield{author}{\bibinfo{person}{Chin-Yew Lin}.}
  \bibinfo{year}{2004}\natexlab{}.
\newblock \showarticletitle{Rouge: A package for automatic evaluation of
  summaries}.
\newblock \bibinfo{journal}{\emph{Text Summarization Branches Out}}
  (\bibinfo{year}{2004}).
\newblock


\bibitem[\protect\citeauthoryear{Lin, Feng, Santos, Yu, Xiang, Zhou, and
  Bengio}{Lin et~al\mbox{.}}{2017}]%
        {lin2017structured}
\bibfield{author}{\bibinfo{person}{Zhouhan Lin}, \bibinfo{person}{Minwei Feng},
  \bibinfo{person}{Cicero Nogueira~dos Santos}, \bibinfo{person}{Mo Yu},
  \bibinfo{person}{Bing Xiang}, \bibinfo{person}{Bowen Zhou}, {and}
  \bibinfo{person}{Yoshua Bengio}.} \bibinfo{year}{2017}\natexlab{}.
\newblock \showarticletitle{A structured self-attentive sentence embedding}.
\newblock \bibinfo{journal}{\emph{arXiv}} (\bibinfo{year}{2017}).
\newblock


\bibitem[\protect\citeauthoryear{Liu, Lowe, Serban, Noseworthy, Charlin, and
  Pineau}{Liu et~al\mbox{.}}{2016}]%
        {liu2016not}
\bibfield{author}{\bibinfo{person}{Chia-Wei Liu}, \bibinfo{person}{Ryan Lowe},
  \bibinfo{person}{Iulian~V Serban}, \bibinfo{person}{Michael Noseworthy},
  \bibinfo{person}{Laurent Charlin}, {and} \bibinfo{person}{Joelle Pineau}.}
  \bibinfo{year}{2016}\natexlab{}.
\newblock \showarticletitle{How not to evaluate your dialogue system: An
  empirical study of unsupervised evaluation metrics for dialogue response
  generation}. In \bibinfo{booktitle}{\emph{EMNLP}}.
\newblock


\bibitem[\protect\citeauthoryear{Luong, Pham, and Manning}{Luong
  et~al\mbox{.}}{2015}]%
        {luong2015effective}
\bibfield{author}{\bibinfo{person}{Minh-Thang Luong}, \bibinfo{person}{Hieu
  Pham}, {and} \bibinfo{person}{Christopher~D Manning}.}
  \bibinfo{year}{2015}\natexlab{}.
\newblock \showarticletitle{Effective approaches to attention-based neural
  machine translation}. In \bibinfo{booktitle}{\emph{EMNLP}}.
\newblock


\bibitem[\protect\citeauthoryear{Martinovski and Traum}{Martinovski and
  Traum}{2003}]%
        {martinovski2003breakdown}
\bibfield{author}{\bibinfo{person}{Bilyana Martinovski} {and}
  \bibinfo{person}{David Traum}.} \bibinfo{year}{2003}\natexlab{}.
\newblock \showarticletitle{Breakdown in human-machine interaction: the error
  is the clue}. In \bibinfo{booktitle}{\emph{ISCA tutorial and research
  workshop}}. \bibinfo{pages}{11--16}.
\newblock


\bibitem[\protect\citeauthoryear{Mikolov, Chen, Corrado, and Dean}{Mikolov
  et~al\mbox{.}}{2013}]%
        {mikolov2013efficient}
\bibfield{author}{\bibinfo{person}{Tomas Mikolov}, \bibinfo{person}{Kai Chen},
  \bibinfo{person}{Greg Corrado}, {and} \bibinfo{person}{Jeffrey Dean}.}
  \bibinfo{year}{2013}\natexlab{}.
\newblock \showarticletitle{Efficient estimation of word representations in
  vector space}.
\newblock \bibinfo{journal}{\emph{arXiv}} (\bibinfo{year}{2013}).
\newblock


\bibitem[\protect\citeauthoryear{Mou, Song, Yan, Li, Zhang, and Jin}{Mou
  et~al\mbox{.}}{2016}]%
        {mou2016sequence}
\bibfield{author}{\bibinfo{person}{Lili Mou}, \bibinfo{person}{Yiping Song},
  \bibinfo{person}{Rui Yan}, \bibinfo{person}{Ge Li}, \bibinfo{person}{Lu
  Zhang}, {and} \bibinfo{person}{Zhi Jin}.} \bibinfo{year}{2016}\natexlab{}.
\newblock \showarticletitle{Sequence to backward and forward sequences: A
  content-introducing approach to generative short-text conversation}. In
  \bibinfo{booktitle}{\emph{COLING}}.
\newblock


\bibitem[\protect\citeauthoryear{Papineni, Roukos, Ward, and Zhu}{Papineni
  et~al\mbox{.}}{2002}]%
        {papineni2002bleu}
\bibfield{author}{\bibinfo{person}{Kishore Papineni}, \bibinfo{person}{Salim
  Roukos}, \bibinfo{person}{Todd Ward}, {and} \bibinfo{person}{Wei-Jing Zhu}.}
  \bibinfo{year}{2002}\natexlab{}.
\newblock \showarticletitle{BLEU: a method for automatic evaluation of machine
  translation}. In \bibinfo{booktitle}{\emph{ACL}}. \bibinfo{pages}{311--318}.
\newblock


\bibitem[\protect\citeauthoryear{Peng, Fang, Xie, and Zhou}{Peng
  et~al\mbox{.}}{2019}]%
        {peng2019topic}
\bibfield{author}{\bibinfo{person}{Yehong Peng}, \bibinfo{person}{Yizhen Fang},
  \bibinfo{person}{Zhiwen Xie}, {and} \bibinfo{person}{Guangyou Zhou}.}
  \bibinfo{year}{2019}\natexlab{}.
\newblock \showarticletitle{Topic-enhanced emotional conversation generation
  with attention mechanism}.
\newblock \bibinfo{journal}{\emph{KBS}}  \bibinfo{volume}{163}
  (\bibinfo{year}{2019}), \bibinfo{pages}{429--437}.
\newblock


\bibitem[\protect\citeauthoryear{Plutchik}{Plutchik}{1980}]%
        {plutchik1980general}
\bibfield{author}{\bibinfo{person}{Robert Plutchik}.}
  \bibinfo{year}{1980}\natexlab{}.
\newblock \showarticletitle{A general psychoevolutionary theory of emotion}.
\newblock In \bibinfo{booktitle}{\emph{Theories of emotion}}.
  \bibinfo{publisher}{Elsevier}, \bibinfo{pages}{3--33}.
\newblock


\bibitem[\protect\citeauthoryear{Plutchik}{Plutchik}{2001}]%
        {plutchik2001nature}
\bibfield{author}{\bibinfo{person}{Robert Plutchik}.}
  \bibinfo{year}{2001}\natexlab{}.
\newblock \showarticletitle{The nature of emotions: Human emotions have deep
  evolutionary roots, a fact that may explain their complexity and provide
  tools for clinical practice}.
\newblock \bibinfo{journal}{\emph{American scientist}} \bibinfo{volume}{89},
  \bibinfo{number}{4} (\bibinfo{year}{2001}), \bibinfo{pages}{344--350}.
\newblock


\bibitem[\protect\citeauthoryear{Prendinger, Mori, and Ishizuka}{Prendinger
  et~al\mbox{.}}{2005}]%
        {prendinger2005using}
\bibfield{author}{\bibinfo{person}{Helmut Prendinger},
  \bibinfo{person}{Junichiro Mori}, {and} \bibinfo{person}{Mitsuru Ishizuka}.}
  \bibinfo{year}{2005}\natexlab{}.
\newblock \showarticletitle{Using human physiology to evaluate subtle
  expressivity of a virtual quizmaster in a mathematical game}.
\newblock \bibinfo{journal}{\emph{IJHCS}} (\bibinfo{year}{2005}),
  \bibinfo{pages}{231--245}.
\newblock


\bibitem[\protect\citeauthoryear{Qian, Huang, Zhao, Xu, and Zhu}{Qian
  et~al\mbox{.}}{2017}]%
        {qian2017assigning}
\bibfield{author}{\bibinfo{person}{Qiao Qian}, \bibinfo{person}{Minlie Huang},
  \bibinfo{person}{Haizhou Zhao}, \bibinfo{person}{Jingfang Xu}, {and}
  \bibinfo{person}{Xiaoyan Zhu}.} \bibinfo{year}{2017}\natexlab{}.
\newblock \showarticletitle{Assigning personality/identity to a chatting
  machine for coherent conversation generation}.
\newblock \bibinfo{journal}{\emph{arXiv}} (\bibinfo{year}{2017}).
\newblock


\bibitem[\protect\citeauthoryear{Ritter, Cherry, and Dolan}{Ritter
  et~al\mbox{.}}{2011}]%
        {ritter2011data}
\bibfield{author}{\bibinfo{person}{Alan Ritter}, \bibinfo{person}{Colin
  Cherry}, {and} \bibinfo{person}{William~B Dolan}.}
  \bibinfo{year}{2011}\natexlab{}.
\newblock \showarticletitle{Data-driven response generation in social media}.
  In \bibinfo{booktitle}{\emph{EMNLP}}. \bibinfo{pages}{583--593}.
\newblock


\bibitem[\protect\citeauthoryear{Serban, Sordoni, Bengio, Courville, and
  Pineau}{Serban et~al\mbox{.}}{2016}]%
        {serban2016building}
\bibfield{author}{\bibinfo{person}{Iulian~Vlad Serban},
  \bibinfo{person}{Alessandro Sordoni}, \bibinfo{person}{Yoshua Bengio},
  \bibinfo{person}{Aaron~C Courville}, {and} \bibinfo{person}{Joelle Pineau}.}
  \bibinfo{year}{2016}\natexlab{}.
\newblock \showarticletitle{Building End-To-End Dialogue Systems Using
  Generative Hierarchical Neural Network Models.}. In
  \bibinfo{booktitle}{\emph{AAAI}}. \bibinfo{pages}{3776--3784}.
\newblock


\bibitem[\protect\citeauthoryear{Serban, Sordoni, Lowe, Charlin, Pineau,
  Courville, and Bengio}{Serban et~al\mbox{.}}{2017}]%
        {serban2017hierarchical}
\bibfield{author}{\bibinfo{person}{Iulian~Vlad Serban},
  \bibinfo{person}{Alessandro Sordoni}, \bibinfo{person}{Ryan Lowe},
  \bibinfo{person}{Laurent Charlin}, \bibinfo{person}{Joelle Pineau},
  \bibinfo{person}{Aaron Courville}, {and} \bibinfo{person}{Yoshua Bengio}.}
  \bibinfo{year}{2017}\natexlab{}.
\newblock \showarticletitle{A hierarchical latent variable encoder-decoder
  model for generating dialogues}. In \bibinfo{booktitle}{\emph{AAAI}}.
\newblock


\bibitem[\protect\citeauthoryear{Shang, Lu, and Li}{Shang
  et~al\mbox{.}}{2015}]%
        {shang2015neural}
\bibfield{author}{\bibinfo{person}{Lifeng Shang}, \bibinfo{person}{Zhengdong
  Lu}, {and} \bibinfo{person}{Hang Li}.} \bibinfo{year}{2015}\natexlab{}.
\newblock \showarticletitle{Neural responding machine for short-text
  conversation}. In \bibinfo{booktitle}{\emph{ACL}}.
\newblock


\bibitem[\protect\citeauthoryear{Song, Shi, Li, and Zhang}{Song
  et~al\mbox{.}}{2018}]%
        {song2018directional}
\bibfield{author}{\bibinfo{person}{Yan Song}, \bibinfo{person}{Shuming Shi},
  \bibinfo{person}{Jing Li}, {and} \bibinfo{person}{Haisong Zhang}.}
  \bibinfo{year}{2018}\natexlab{}.
\newblock \showarticletitle{Directional Skip-Gram: Explicitly Distinguishing
  Left and Right Context for Word Embeddings}. In
  \bibinfo{booktitle}{\emph{NAACL}}. \bibinfo{pages}{175--180}.
\newblock


\bibitem[\protect\citeauthoryear{Sordoni, Galley, Auli, Brockett, Ji, Mitchell,
  Nie, Gao, and Dolan}{Sordoni et~al\mbox{.}}{2015}]%
        {sordoni2015neural}
\bibfield{author}{\bibinfo{person}{Alessandro Sordoni}, \bibinfo{person}{Michel
  Galley}, \bibinfo{person}{Michael Auli}, \bibinfo{person}{Chris Brockett},
  \bibinfo{person}{Yangfeng Ji}, \bibinfo{person}{Margaret Mitchell},
  \bibinfo{person}{Jian-Yun Nie}, \bibinfo{person}{Jianfeng Gao}, {and}
  \bibinfo{person}{Bill Dolan}.} \bibinfo{year}{2015}\natexlab{}.
\newblock \showarticletitle{A neural network approach to context-sensitive
  generation of conversational responses}.
\newblock \bibinfo{journal}{\emph{arXiv}} (\bibinfo{year}{2015}).
\newblock


\bibitem[\protect\citeauthoryear{Sutskever, Vinyals, and Le}{Sutskever
  et~al\mbox{.}}{2014}]%
        {sutskever2014sequence}
\bibfield{author}{\bibinfo{person}{Ilya Sutskever}, \bibinfo{person}{Oriol
  Vinyals}, {and} \bibinfo{person}{Quoc~V Le}.}
  \bibinfo{year}{2014}\natexlab{}.
\newblock \showarticletitle{Sequence to sequence learning with neural
  networks}. In \bibinfo{booktitle}{\emph{NIPS}}. \bibinfo{pages}{3104--3112}.
\newblock


\bibitem[\protect\citeauthoryear{Tang, Wei, Yang, Zhou, Liu, and Qin}{Tang
  et~al\mbox{.}}{2014}]%
        {tang2014learning}
\bibfield{author}{\bibinfo{person}{Duyu Tang}, \bibinfo{person}{Furu Wei},
  \bibinfo{person}{Nan Yang}, \bibinfo{person}{Ming Zhou},
  \bibinfo{person}{Ting Liu}, {and} \bibinfo{person}{Bing Qin}.}
  \bibinfo{year}{2014}\natexlab{}.
\newblock \showarticletitle{Learning sentiment-specific word embedding for
  twitter sentiment classification}. In \bibinfo{booktitle}{\emph{ACL}}.
  \bibinfo{pages}{1555--1565}.
\newblock


\bibitem[\protect\citeauthoryear{Vaswani, Shazeer, Parmar, Uszkoreit, Jones,
  Gomez, Kaiser, and Polosukhin}{Vaswani et~al\mbox{.}}{2017}]%
        {vaswani2017attention}
\bibfield{author}{\bibinfo{person}{Ashish Vaswani}, \bibinfo{person}{Noam
  Shazeer}, \bibinfo{person}{Niki Parmar}, \bibinfo{person}{Jakob Uszkoreit},
  \bibinfo{person}{Llion Jones}, \bibinfo{person}{Aidan~N Gomez},
  \bibinfo{person}{{\L}ukasz Kaiser}, {and} \bibinfo{person}{Illia
  Polosukhin}.} \bibinfo{year}{2017}\natexlab{}.
\newblock \showarticletitle{Attention is all you need}. In
  \bibinfo{booktitle}{\emph{NIPS}}. \bibinfo{pages}{5998--6008}.
\newblock


\bibitem[\protect\citeauthoryear{Vijayakumar, Cogswell, Selvaraju, Sun, Lee,
  Crandall, and Batra}{Vijayakumar et~al\mbox{.}}{2018}]%
        {vijayakumar2016diverse}
\bibfield{author}{\bibinfo{person}{Ashwin~K Vijayakumar},
  \bibinfo{person}{Michael Cogswell}, \bibinfo{person}{Ramprasath~R Selvaraju},
  \bibinfo{person}{Qing Sun}, \bibinfo{person}{Stefan Lee},
  \bibinfo{person}{David Crandall}, {and} \bibinfo{person}{Dhruv Batra}.}
  \bibinfo{year}{2018}\natexlab{}.
\newblock \showarticletitle{Diverse beam search: Decoding diverse solutions
  from neural sequence models}. In \bibinfo{booktitle}{\emph{AAAI}}.
\newblock


\bibitem[\protect\citeauthoryear{Vinyals, Kaiser, Koo, Petrov, Sutskever, and
  Hinton}{Vinyals et~al\mbox{.}}{2015}]%
        {vinyals2015grammar}
\bibfield{author}{\bibinfo{person}{Oriol Vinyals}, \bibinfo{person}{{\L}ukasz
  Kaiser}, \bibinfo{person}{Terry Koo}, \bibinfo{person}{Slav Petrov},
  \bibinfo{person}{Ilya Sutskever}, {and} \bibinfo{person}{Geoffrey Hinton}.}
  \bibinfo{year}{2015}\natexlab{}.
\newblock \showarticletitle{Grammar as a foreign language}. In
  \bibinfo{booktitle}{\emph{NIPS}}. \bibinfo{pages}{2773--2781}.
\newblock


\bibitem[\protect\citeauthoryear{Vinyals and Le}{Vinyals and Le}{2015}]%
        {vinyals2015neural}
\bibfield{author}{\bibinfo{person}{Oriol Vinyals} {and} \bibinfo{person}{Quoc
  Le}.} \bibinfo{year}{2015}\natexlab{}.
\newblock \showarticletitle{A neural conversational model}. In
  \bibinfo{booktitle}{\emph{ICML Deep Learning Workshop}}.
\newblock


\bibitem[\protect\citeauthoryear{Wallace}{Wallace}{2003}]%
        {wallace2003elements}
\bibfield{author}{\bibinfo{person}{Richard Wallace}.}
  \bibinfo{year}{2003}\natexlab{}.
\newblock \showarticletitle{The elements of AIML style}.
\newblock \bibinfo{journal}{\emph{Alice AI Foundation}} (\bibinfo{year}{2003}).
\newblock


\bibitem[\protect\citeauthoryear{Wilcox}{Wilcox}{2011}]%
        {wilcox2011beyond}
\bibfield{author}{\bibinfo{person}{Bruce Wilcox}.}
  \bibinfo{year}{2011}\natexlab{}.
\newblock \showarticletitle{Beyond Fa{\c{c}}ade: Pattern matching for natural
  language applications}.
\newblock \bibinfo{journal}{\emph{GamaSutra. com}} (\bibinfo{year}{2011}).
\newblock


\bibitem[\protect\citeauthoryear{Xing, Wu, Wu, Liu, Huang, Zhou, and Ma}{Xing
  et~al\mbox{.}}{2017}]%
        {xing2017topic}
\bibfield{author}{\bibinfo{person}{Chen Xing}, \bibinfo{person}{Wei Wu},
  \bibinfo{person}{Yu Wu}, \bibinfo{person}{Jie Liu}, \bibinfo{person}{Yalou
  Huang}, \bibinfo{person}{Ming Zhou}, {and} \bibinfo{person}{Wei-Ying Ma}.}
  \bibinfo{year}{2017}\natexlab{}.
\newblock \showarticletitle{Topic Aware Neural Response Generation.}. In
  \bibinfo{booktitle}{\emph{AAAI}}. \bibinfo{pages}{3351--3357}.
\newblock


\bibitem[\protect\citeauthoryear{Zhang, Lan, Guo, Xu, and Cheng}{Zhang
  et~al\mbox{.}}{2018}]%
        {zhang2018tailored}
\bibfield{author}{\bibinfo{person}{Hainan Zhang}, \bibinfo{person}{Yanyan Lan},
  \bibinfo{person}{Jiafeng Guo}, \bibinfo{person}{Jun Xu}, {and}
  \bibinfo{person}{Xueqi Cheng}.} \bibinfo{year}{2018}\natexlab{}.
\newblock \showarticletitle{Tailored Sequence to Sequence Models to Different
  Conversation Scenarios}. In \bibinfo{booktitle}{\emph{ACL}}.
  \bibinfo{pages}{1479--1488}.
\newblock


\bibitem[\protect\citeauthoryear{Zhou, Huang, Zhang, Zhu, and Liu}{Zhou
  et~al\mbox{.}}{2018a}]%
        {zhou2018emotional}
\bibfield{author}{\bibinfo{person}{Hao Zhou}, \bibinfo{person}{Minlie Huang},
  \bibinfo{person}{Tianyang Zhang}, \bibinfo{person}{Xiaoyan Zhu}, {and}
  \bibinfo{person}{Bing Liu}.} \bibinfo{year}{2018}\natexlab{a}.
\newblock \showarticletitle{Emotional chatting machine: Emotional conversation
  generation with internal and external memory}. In
  \bibinfo{booktitle}{\emph{AAAI}}.
\newblock


\bibitem[\protect\citeauthoryear{Zhou, Young, Huang, Zhao, Xu, and Zhu}{Zhou
  et~al\mbox{.}}{2018b}]%
        {zhou2018commonsense}
\bibfield{author}{\bibinfo{person}{Hao Zhou}, \bibinfo{person}{Tom Young},
  \bibinfo{person}{Minlie Huang}, \bibinfo{person}{Haizhou Zhao},
  \bibinfo{person}{Jingfang Xu}, {and} \bibinfo{person}{Xiaoyan Zhu}.}
  \bibinfo{year}{2018}\natexlab{b}.
\newblock \showarticletitle{Commonsense Knowledge Aware Conversation Generation
  with Graph Attention.}. In \bibinfo{booktitle}{\emph{IJCAI}}.
  \bibinfo{pages}{4623--4629}.
\newblock


\bibitem[\protect\citeauthoryear{Zhou and Wang}{Zhou and Wang}{2017}]%
        {zhou2017mojitalk}
\bibfield{author}{\bibinfo{person}{Xianda Zhou} {and}
  \bibinfo{person}{William~Yang Wang}.} \bibinfo{year}{2017}\natexlab{}.
\newblock \showarticletitle{Mojitalk: Generating emotional responses at scale}.
\newblock \bibinfo{journal}{\emph{arXiv}} (\bibinfo{year}{2017}).
\newblock


\end{thebibliography}

\end{document}